\documentclass{article}

\usepackage[preprint,nonatbib]{nips_2018}
\usepackage{nips_2018}
\usepackage{bm}
\usepackage[utf8]{inputenc}
\usepackage{url}
\usepackage{amssymb}
\usepackage{amsfonts}
\usepackage{graphicx}
\usepackage{url}
\usepackage{booktabs}
\usepackage{xspace}
\usepackage{lstsemantic}
\usepackage{lstlangmizar}

\usepackage{tikz}
\usetikzlibrary{positioning}
\usetikzlibrary{shapes.geometric,arrows.meta,decorations.markings}
\usetikzlibrary{calc,shapes.callouts}

\usepackage{float}
\usepackage{hyperref}

\def\systemname#1{\textsf{#1}\xspace}

\newcommand{\rlc}{\systemname{rlCoP}}
\newcommand{\mlc}{\systemname{mlCoP}}
\newcommand{\lc}{\systemname{leanCoP}}

\title{Reinforcement Learning of Theorem Proving}

\author{Cezary Kaliszyk\thanks{These authors contributed equally to this work.} \thanks{Supported by ERC grant no.\ 714034 \textit{SMART}.} \\
University of Innsbruck
  \And Josef Urban \footnotemark[1]   \thanks{Supported by the \textit{AI4REASON} ERC Consolidator grant number 649043, and by the Czech project AI\&Reasoning CZ.02.1.01/0.0/0.0/15\_003/0000466 and the European Regional Development Fund.} \\
Czech Technical University in Prague
\AND Henryk Michalewski \\
University of Warsaw
\And Mirek Ol\v{s}\'{a}k \\
Charles University
}

\begin{document}

\maketitle

\begin{abstract}
  We introduce a theorem proving algorithm that uses practically no
  domain heuristics for guiding 
its connection-style proof
  search. Instead, it runs many Monte-Carlo simulations guided by
  reinforcement learning from previous proof attempts.  
We produce
   several versions of the prover, parameterized by different
   learning and guiding algorithms. 
The strongest
  version of the system is trained %
on a
  large corpus of mathematical problems and evaluated on previously unseen
  problems. The trained system solves 
within the same number of inferences
over 40\% more problems
  than a baseline %
prover,  
which is an unusually high improvement in this hard AI domain. To our knowledge this is
the first time reinforcement learning has been convincingly applied to solving general mathematical problems on a large scale.
\end{abstract}

\section{Introduction}
\label{Introduction}
Automated theorem proving (ATP)~\cite{DBLP:books/el/RobinsonV01} can
in principle
be
used
to
 attack
any formally stated mathematical problem.
For this, state-of-the-art ATP systems rely on fast implementations of
complete proof calculi such as resolution~\cite{robinson1965machine},
superposition~\cite{bachmair1994rewrite},
SMT~\cite{barrett2009satisfiability} and (connection)
tableau~\cite{DBLP:books/el/RV01/Hahnle01} that have been over several
decades improved by many search heuristics. This is already useful for
automatically discharging smaller proof obligations in large
interactive theorem proving (ITP) verification projects~\cite{hammers4qed}.  In
practice, today's best ATP system are however still far weaker than
trained mathematicians in most research domains. Machine learning from
many proofs could be used to improve on this.

Following this idea, large formal proof corpora have been recently
translated to ATP formalisms~\cite{Urban06,MengP08,holyhammer},
and machine learning over them has started to be used to train
guidance of ATP systems~\cite{US+08-long,KuhlweinLTUH12-long,IrvingSAECU16}. First, to select a small
number of relevant facts for proving new conjectures over large formal
libraries~\cite{abs-1108-3446,BlanchetteGKKU16,hh4h4}, and more
recently also to guide the internal search of the ATP systems. In
sophisticated saturation-style provers this has been done by feedback
loops for strategy invention~\cite{blistr,JakubuvU17,SchaferS15} and
by using supervised learning~\cite{JakubuvU17a,LoosISK17} to select
the next given clause~\cite{mccune1990otter}. In the simpler connection
tableau systems such as \lc~\cite{OB03}, supervised learning has
been used to choose the next tableau extension
step~\cite{UrbanVS11,KaliszykU15} and first experiments with
Monte-Carlo guided proof search~\cite{FarberKU17} have been done.
Despite a limited ability to prioritize the proof search, the guided
search in the latter connection systems is still organized by
\emph{iterative deepening}. This ensures completeness, which has been
for long time a \emph{sine qua non} for building proof calculi.

In this work, we remove this requirement, since it basically means
that all shorter proof candidates have to be tried before a longer
proof is found.  The result is a bare connection-style theorem prover
that does not use any human-designed proof-search restrictions,
heuristics and targeted (decision) procedures.  This is in stark
contrast to recent mainstream ATP research, which has to a large
extent focused on adding more and more sophisticated human-designed
procedures in domains such as SMT solving.

Based on the bare prover, we build a sequence of systems, adding
Monte-Carlo tree search~\cite{DBLP:conf/ecml/KocsisS06}, and reinforcement learning~\cite{sutton1998reinforcement} of policy and value guidance. We show that while the performance of the system (called \rlc) is initially
much worse than that of standard \lc, after ten iterations of proving and learning it
solves significantly more 
previously unseen
  problems than \lc when using the same total number of
inference steps.

The rest of the paper is organized as follows. Section~\ref{Prover}
explains the basic connection tableau setting and introduces the bare
prover. Section~\ref{Guidance} describes integration of the
learning-based guiding mechanisms, i.e. Monte-Carlo search, the policy
and value guidance.
Section~\ref{Experiments} evaluates the
system on a large corpus of problems extracted from the Mizar
Mathematical Library~\cite{mizar-in-a-nutshell}.

\section{The Game of Connection Based Theorem Proving}
\label{Prover}

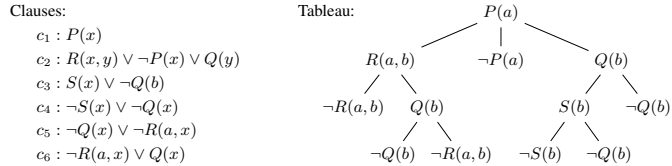
\begin{figure}[thb]
\begin{center}
\scalebox{.7}{
  \small
  \begin{minipage}{.35\textwidth}
  \hspace{-5mm}Clauses:\\[1mm]
  $c_1 : P(x)$\\[1mm]
  $c_2 : R(x,y)\lor\lnot P(x) \lor Q(y)$\\[1mm]
  $c_3 : S(x)\lor\lnot Q(b)$\\[1mm]
  $c_4 : \lnot S(x)\lor \lnot Q(x)$\\[1mm]
  $c_5 : \lnot Q(x)\lor \lnot R(a,x)$\\[1mm]
  $c_6 : \lnot R(a,x) \lor Q(x)$
  \end{minipage}
  \raisebox{13.8mm}{Tableau:}\hspace{-8mm}
  \begin{minipage}{.45\textwidth}
\begin{tikzpicture}[level distance=9mm,
  level 1/.style ={sibling distance=2.1cm},
  level 2/.style ={sibling distance=1.4cm},
  level 3/.style ={sibling distance=1.3cm},
  ]
\node{$P(a)$}
  child { node {$R(a,b)$}
    child { node {$\lnot R(a,b)$} }
    child { node {$Q(b)$}
      child { node {$\lnot Q(b)$}}
      child { node {$\lnot R(a,b)$}}
    }
  }
  child { node {$\lnot P(a)$} }
  child { node {$Q(b)$}
    child { node {$S(b)$}
      child { node {$\lnot S(b)$}}
      child { node {$\lnot Q(b)$}}
    }
    child { node {$\lnot Q(b)$}}
  };
\end{tikzpicture}
\end{minipage}}
\end{center}
\caption{\label{tableau}Closed connection tableau for a set of clauses (adapted from Letz et al.~\cite{Letz1994ControlledIO}).}
\end{figure}

We assume basic first-order logic and
theorem proving terminology~\cite{DBLP:books/el/RobinsonV01}.
We start with the connection tableau architecture as implemented by
the \lc~\cite{OB03} system. \lc is a compact theorem prover
whose core procedure can be written in seven lines in Prolog. Its
input is a (mathematical) problem consisting of \emph{axioms} and
\emph{conjecture} formally stated in first-order logic (FOL).
The calculus searches for \emph{refutational proofs},
i.e. proofs showing that
the axioms together with the negated conjecture are
\emph{unsatisfiable}.\footnote{To minimize the required theorem proving background, we follow the more standard connection tableau calculus presentations using CNF and refutational setting as in~\cite{Letz1994ControlledIO}. The \lc calculus is typically presented in a dual (DNF) form, which is however isomorphic to the more standard one.}   
The FOL formulas are first translated to the
\emph{clause normal form} (CNF), producing a set of first-order
\emph{clauses} consisting of \emph{literals} (atoms or their
negations).  An example set of clauses is shown in
Figure~\ref{tableau}. The figure also shows a \emph{closed connection
  tableau}, i.e., a finished proof tree where every branch contains
\emph{complementary literals} (literals with opposite polarity). Since all branches contain a pair of contradictory literals, this shows that the set of clauses is unsatisfiable.

The proof search starts with a \emph{start clause} as a \emph{goal} and
proceeds by building a connection tableau by repeatedly
applying \emph{extension steps} and \emph{reduction steps} to it. 
The
extension step connects (\emph{unifies}) the \emph{current goal} (a selected tip of a tableau branch) with a
complementary literal of a new clause. This extends the
\emph{current branch}, possibly splitting it into several branches if there are more literals in the new clause,
and possibly \emph{instantiating} some variables in the tableau. The
reduction step connects the current goal to a complementary literal of
the \emph{active path}, thus \emph{closing} the current branch. The
proof is finished when all branches are closed. The extension and
reduction steps are nondeterministic, requiring backtracking in the
standard connection calculus. \emph{Iterative deepening} is typically
used to ensure completeness, i.e. making sure that the proof search
finds a proof if there is any. \emph{Incomplete strategies} that restrict backtracking
can be used in \lc, sometimes improving its performance on benchmarks~\cite{DBLP:journals/aicom/Otten10}.

\subsection{The Bare Prover}
Our bare prover is based on a previous
reimplementation~\cite{ckjujv-cpp15} of \lc in OCaml
(\emph{\mlc}). Unlike the Prolog version, \mlc uses an explicit
stack for storing the full proof state. This allows us to use the full
proof state for machine learning guidance.  We first modify \mlc by
removing iterative deepening, i.e., the traversal strategy that makes
sure that shorter (shallower) tableaux are tested before deeper
ones. Instead, the bare prover randomly chooses extension and
reduction steps operating on the current goal, possibly going into
arbitrary depth. %
This makes our bare prover trivially incomplete. A simple example to
demonstrate that is the unsatisfiable set of clauses
$\{P(0), \lnot P(x) \lor P(s(x)), \lnot P(0)\}$ . If the prover starts
with the goal $P(0)$, the third clause can be used to immediately
close the tableau. Without any depth bounds, the bare prover may
however also always extend with the second clause, generating an
infinite branch $P(0), P(s(0)), P(s(s(0))), ... $.
Rather than designing such completeness bounds and corresponding
exhaustive strategies, we will use Monte-Carlo search and
reinforcement learning to gradually teach the prover to avoid such bad
branches and focus on the promising ones.

Next, we add playouts and search node visit counts. A \emph{playout}
of length $d$ is simply a sequence of $d$ consecutive
extension/reduction steps (\emph{inferences}) from a given
\emph{proof state} (a tableau with a selected goal).  Inferences thus
correspond to \emph{actions} and are similar to moves in games.  We represent
inferences as integers that encode the selected clause together with the literal
that connected to the goal.  Instead of running one potentially
infinite playout, the bare prover can be instructed to play $n$
playouts of length $d$. Each playout updates the counts for the \emph{search
nodes} that it visits. Search nodes are encoded as
sequences of inferences starting at the empty tableau.
A playout can also run without length restrictions until it visits a previously unexplored search node. 
This is the current default. Each
playout in this version always starts with empty tableau, i.e., it
starts randomly from scratch.

The next modification to this simple setup are \emph{bigsteps} done
after $b$ playouts. They correspond to moves that are chosen in games
after many playouts. Similarly, instead of starting all playouts always from
scratch (empty tableau) as above, we choose after the first $b$ playouts a particular single
inference (bigstep), %
resulting in a new \emph{bigstep tableau}. The next $b$ playouts
will start with this tableau, followed by another bigstep, etc.

This finishes the description of the bare prover. Without any guidance
and heuristics for choosing the playout inferences and bigsteps, this
prover is typically much weaker than standard \mlc with iterative
deepening, see Section~\ref{Experiments}.  The bare prover will just iterate between randomly doing
$b$ new playouts, and randomly making a bigstep.

\section{Guidance}
\label{Guidance}
The \rlc extends the bare prover with 
(i) Monte-Carlo tree search balancing exploration and exploitation using the UCT formula~\cite{DBLP:conf/ecml/KocsisS06},
(ii) learning-based mechanisms for estimating the prior probability of inferences to lead to a proof (\emph{policy}), and 
(iii) learning-based mechanisms for assigning heuristic \emph{value} to the proof states (tableaux).

\subsection{Monte-Carlo Guidance}

To implement Monte-Carlo tree search, we maintain at each search node
$i$ the number of its visits $n_i$, the total reward $w_i$, and its
prior probability $p_i$. This is the transition probability of the
action (inference) that leads from $i$'s  parent node to $i$.
If no policy learning is used, the prior
probabilities are all equal to one.
The total reward for a node is computed as a sum of
the rewards of all nodes below that node.  In the basic setting, the reward
for a leaf node is $1$ if the sequence of inferences results in a
closed tableau, i.e., a proof of the conjecture. Otherwise it is $0$. 

Instead of this basic setting, we will by default use a simple evaluation
heuristic, that will later be replaced by learned value. The heuristic is
based on the number of open (non-closed) goals (tips of the tableau) $G_o$. The exact
value is computed as $0.95^{G_o}$, i.e., the leaf value exponentially
drops with the number of open goals in the tableau. The motivation is
similar as, e.g., preferring smaller clauses (closer to the empty
clause) in saturation-style theorem provers. If nothing else is known
and the open goals are assumed to be independent, the chances of
closing the tableau within a given inference limit drop exponentially
with each added open goal. The exact value of $0.95$ has been determined
experimentally using a small grid search.

We use the standard UCT formula~\cite{DBLP:conf/ecml/KocsisS06} to select the next inferences (actions) in the playouts:
    \[\frac{w_i}{n_i} + c\cdot p_i \cdot \sqrt{\frac{\ln{{N}}}{ n_i}} \]
    where $N$ stands for the total number of visits of the parent
    node.  We have also experimented with PUCT as in
    AlphaZero~\cite{silver2017mastering}, however the results are
    practically the same. The value of $c$ has been experimentally set to $2$ when learned policy and value are used.

\subsection{Policy Learning and Guidance}
\label{sec:policy}

From many proof runs we learn prior probabilities of actions
(inferences) in particular proof states corresponding to the search
nodes in which the actions were taken.  We characterize the proof
states for policy learning by extracting features (see
Section~\ref{Features}) from the current goal, the active path, and
the whole tableau. Similarly, we extract features from the clause and
its literal that were used to perform the inference. Both are
extracted as sparse vectors and concatenated into pairs
$(f_{state},f_{action})$. For each search node, we extract from its
UCT data the frequency of each action $a$, and normalize it by
dividing with the average action frequency at that node. This yields a
relative proportion $r_a \in (0, \infty)$.  Each concatenated pair of
feature vectors $(f_{s},f_{a})$ is then associated with $r_a$, which
constitutes the training data for policy learning implemented as
regression on the logarithms. During the proof search, the prior
probabilities $p_i$ of the available actions $a_i$ in a state $s$ are
computed as a softmax of their predictions. We use $\tau = 2.5$ by
default as the softmax temperature. This value has been optimized by a
small grid search.

The policy learning data can be extracted from all search nodes or
only from some of them. By default, we only extract the training
examples from the bigstep nodes. This makes the amount of training
data manageable for our experiments and also focuses on important
examples.

\subsection{Value Learning and Guidance}
\label{sec:value}
Bigstep nodes are used also for learning of the proof state evaluation
(value).  For value learning  we characterize the  proof states of the nodes by
extracting features from all goals, the active path, and the whole
tableau. If a proof was found, each bigstep node $b$ is assigned value
$v_b = 1$. If the proof search was unsuccessful, each bigstep is
assigned value $v_b = 0$. By default we also apply a small discount
factor to the positive bigstep values, based on their distance $d_{proof}(b)$ to the closed tableau, measured by the number of inferences. This is computed as
$0.99^{d_{proof}(b)}$. The exact value of $0.99$ has again been
determined experimentally using a small grid search.

For each bigstep node $b$ the sparse vector of its proof state
features $f_{b}$ is associated with the value $v_b$. This constitutes
the training data for value learning which is implemented as
regression on the logits. The trained predictor is then used during the
proof search to estimate the logit of the proof state value.

\subsection{Features}
\label{Features}
We have briefly experimented with using deep neural networks to learn
the policy and value predictors directly from the theorem proving
data.  Current deep neural architectures however do not seem to
perform on such data significantly better than non-neural classifiers
such as XGBoost with manually engineered
features~\cite{JakubuvU17a,abs-1802-03375,IrvingSAECU16}. We use the latter approach, which is also
significantly faster~\cite{LoosISK17}.

Features are collected from the first-order terms, clauses, goals and
tableaux. Most of them are based on (normalized) term walks of length up to $3$, as used in the
ENIGMA system~\cite{JakubuvU17a}. We uniquely identify each symbol by
a 64-bit integer.  To combine a sequence of integers originating from symbols in a term walk
into a single integer, the components are multiplied by fixed large primes and added. The
resulting integers are then reduced to a smaller feature space by taking modulo
by a large prime ($2^{18}-5$). The value of each feature is the sum of
its occurrences in the given expression.

In addition to the term walks we also use several common abstract
features, especially for more complicated data such as tableaux and
paths.  These are: number of goals, total symbol size of all goals,
maximum goal size, maximum goal depth, length of the active path,
number of current variable instantiations, and the two most common
symbols and their frequencies. The exact features used have been
optimized based on several experiments and analysis of the
reinforcement learning data.

\subsection{Learners and Their Integration}
For both policy and value we have experimented with several fast
linear learners such as LIBLINEAR~\cite{fan2008liblinear} and the
XGBoost~\cite{xgboost} gradient boosting toolkit (used with the linear
regression objective). The latter performs significantly better and
has been chosen for conducting the final evaluation. The XGBoost
parameters have been optimized on a smaller dataset using randomized
cross-validated search\footnote{We have used the RandomizedSearchCV
  method of Scikit-learn~\cite{scikit-learn}.}  taking speed of
training and evaluation into account.  The final values that we use
both for policy and value learning are as follows: maximum number of
iterations $= 400$, maximum tree depth $= 9$, ETA (learning rate)
$= 0.3$, early stopping rounds $= 200$, lambda (weight regularization)
$= 1.5$. Table~\ref{MLeval} compares the performance of
XGBoost and LIBLINEAR\footnote{We use L2-regularized L2-loss support vector regression and $\epsilon = 0.0001$ for LIBLINEAR.}
 on value data extracted from 2003 proof attempts.
\begin{table}[htbp]
\begin{small}
  \caption{\label{MLeval} Machine learning performance of XGBoost and LIBLINEAR on the value data extracted from 2003 problems.%
The errors are errors on the logits.}
\centering
\begin{tabular}{lrrr} \toprule
  Predictor & Train Time & RMSE (Train) & RMSE (Test) \\ \midrule
  XGBoost & 19 min & 0.99 & 2.89 \\
  LIBLINEAR & 37 min & 0.83 & 16.31 \\ \bottomrule
\end{tabular}
\end{small}
\end{table}

For real-time guidance during the proof search we have integrated LIBLINEAR and XGBoost into \rlc using the
OCaml foreign interface, which allows for a reasonably low prediction overhead.
Another part of the guidance overhead is feature
computation and transformation of the computed feature vectors into
the form accepted by the learned predictors. Table~\ref{Speed} shows
that the resulting slowdown is in low linear factors.
\begin{table}[htbp]
\begin{small}
  \caption{\label{Speed} Inference speed comparison of \mlc and \rlc. IPS stand for inferences per second. The data are averaged over 2003 problems.}
\centering
  \begin{tabular}{llll}
    \toprule
System &    \mlc &   \rlc without policy/value (UCT only) & \rlc with XGBoost policy/value  \\
Average IPS & 64335.5 & 64772.4 &  16205.7 \\\bottomrule
  \end{tabular}
\end{small}
\end{table}

\section{Experimental Results}
\label{Experiments}

The evaluation is done on two datasets of first-order problems
exported from the Mizar Mathematical
Library~\cite{mizar-in-a-nutshell} by the MPTP system~\cite{Urban06}.
The larger \emph{Miz40}
dataset~\footnote{\url{https://github.com/JUrban/deepmath}} consists
of 32524 problems that have been proved by several state-of-the-art ATPs used with many strategies and high time limits in the experiments described in~\cite{KaliszykU13b}. We have also created a smaller
\emph{M2k} dataset by taking 2003 Miz40 problems that come from
related Mizar articles. Most experiments and tuning were done on the smaller dataset.
All problems are run on the same hardware\footnote{Intel(R) Xeon(R)
  CPU E5-2698 v3 @ 2.30GHz with 256G RAM.} and with the same memory
limits. When using UCT, we always run 2000 playouts (each until a new node is found) per bigstep.

\subsection{Performance without Learning}
\label{sec:nolearn}

First, we use the M2k dataset to compare the performance of the
baseline \mlc with the bare prover and with the non-learning \rlc using
only UCT with the simple goal-counting proof state evaluation
heuristic. The results of runs with a limit of $200000$ inferences are
shown in Table~\ref{NoLearn}. Raising the inference limit helps only a
little: \mlc solves $1003$ problems with a limit of $2*10^6$
inferences, and $1034$ problems with a limit of $4*10^6$ inferences.
The performance of the bare prover is low as expected - only about
half of the performance of \mlc. \rlc using UCT with no policy and only the
simple proof state evaluation heuristic is also weaker than \mlc,
however already significantly better than the bare prover.

\begin{table}[htbp]
\begin{small}
  \caption{\label{NoLearn} Performance on the M2k dataset of \mlc, the bare prover and non-learning \rlc with UCT and simple goal-counting proof state evaluation. (200000 inference limit).}
\centering
  \begin{tabular}{llll}
    \toprule
System &    \mlc &   bare prover & \rlc without policy/value (UCT only) \\
Problems proved & 876 &  434 &  770 \\\bottomrule
  \end{tabular}
\end{small}
\end{table}

\subsection{Reinforcement Learning of Policy Only}
Next we evaluate on the M2k dataset \rlc with UCT using only policy learning, i.e., the
value is still estimated heuristically. We run $20$ iterations, each
with 200000 inference limit. After each iteration we use the policy
training data (Section~\ref{sec:policy}) from all previous iterations
to train a new XGBoost predictor. This is then used for estimating the
prior action probabilities in the next run.  The $0$th run uses no
policy. This means that it is the same as in
Section~\ref{sec:nolearn}, solving 770
problems. Table~\ref{ResultsTablePol} shows the problems solved by
iterations $1$ to $20$.  Already the first iteration significantly
improves over \mlc run with 200000 inference limit. Starting with the
fourth iteration, \rlc is better than \mlc run with the much higher $4*10^6$ inference
limit.

\begin{table}[htbp]
\begin{small}
  \caption{\label{ResultsTablePol}20 policy-guided iterations of \rlc on the M2k dataset.}
\centering
  \begin{tabular}{lllllllllll}
    \toprule
Iteration &   1 &    2 &    3 &    4 &    5 &    6 &    7 &    8 &    9 &   10 \\
Proved & 974 & 1008 & 1028 & 1053 & 1066 & 1054 & 1058 & 1059 & 1075 & 1070 \\\hline
Iteration &   11 &   12 &   13 &   14 &   15 &   16 &   17 &   18 &   19 &   20 \\
Proved & 1074 & 1079 & 1077 & 1080 & 1075 & 1075 & {\bf 1087} & 1071 & 1076 & 1075 \\\bottomrule
  \end{tabular}
\end{small}
\end{table}

\subsection{Reinforcement Learning of Value Only}
Similarly, we evaluate on the M2k dataset $20$ iterations of \rlc with UCT and value learning,
but with no learned policy (i.e., all prior inference probabilities
are the same).  Each iteration again uses a limit of 200000
inferences. After each iteration a new XGBoost predictor is trained on
the value data (Section~\ref{sec:value}) from all previous iterations, and is used to evaluate the
proof states in the next iteration. The $0$th run again uses neither
policy nor value, solving $770$ problems. Table~\ref{ResultsTableVal}
shows the problems solved by iterations $1$ to $20$.  The performance
nearly reaches \mlc, however it is far below \rlc using policy
learning.

\begin{table}[htbp]
\begin{small}
  \caption{\label{ResultsTableVal}20 value-guided iterations of \rlc on the M2k dataset.}
\centering
  \begin{tabular}{lllllllllll}
    \toprule
Iteration&   1 &   2 &   3 &   4 &   5 &   6 &   7 &   8 &   9 &  10 \\
Proved & 809 & 818 & 821 & 821 & 818 & 824 & {\bf 856} & 831 & 842 & 826 \\\hline
Iteration&  11 &  12 &  13 &  14 &  15 &  16 &  17 &  18 &  19 &  20   \\
Proved & 832 & 830 & 825 & 832 & 828 & 820 & 825 & 825 & 831 & 815 \\\bottomrule
  \end{tabular}
\end{small}
\end{table}

\subsection{Reinforcement Learning of Policy and Value}
Finally, we run on the M2k dataset $20$ iterations of full \rlc with
UCT and both policy and value learning.  The inference limits, the
$0$th run and the policy and value learning are as above.
Table~\ref{ResultsTable} shows the problems solved by iterations $1$
to $20$.  The $20$th iteration proves $1235$ problems, which is
$19.4$\% more than \mlc with $4*10^6$ inferences, $13.6$\% more than
the best iteration of \rlc with policy only, and $44.3$\% more than
the best iteration of \rlc with value only.  The first iteration
improves over \mlc with $200000$ inferences by $18.4$\% and the
second iteration already outperforms the best policy-only result.

\begin{table}[htbp]
\begin{small}
  \caption{\label{ResultsTable}20 iterations of \rlc with policy and value guidance on the M2k dataset.}
\centering
  \begin{tabular}{lllllllllll}
    \toprule
Iteration &    1 &    2 &    3 &    4 &    5 &    6 &    7 &    8 &    9 &   10 \\
Proved & 1037 & 1110 & 1166 & 1179 & 1182 & 1198 & 1196 & 1193 & 1212 & 1210 \\\hline
Iteration &   11 &   12 &   13 &   14 &   15 &   16 &   17 &   18 &   19 &   20 \\
Proved & 1206 & 1217 & 1204 & 1219 & 1223 & 1225 & 1224 & 1217 & 1226 & {\bf 1235} \\\bottomrule
  \end{tabular}
\end{small}
\end{table}

We also evaluate the effect of joint reinforcement learning of policy
and value. Replacing the final policy with the best one from the
policy-only runs decreases the performance in 20th iteration from 1235
to 1182. Replacing the final value with the best one from the
value-only runs decreases the performance in 20th iteration from 1235
to 1144.

\subsection{Evaluation on the Whole Miz40 Dataset}
The Miz40 dataset is sufficiently large to allow an ultimate
train/test evaluation in which \rlc is trained in several iterations
on 90\% of the problems, and then compared to \mlc on the 10\% of
previously unseen problems. This will provide the final comparison of
human-designed proof search with proof search trained by reinforcement
learning on many related problems. We therefore randomly split Miz40
into a training set of 29272 problems and a testing set of 3252
problems.

First, we again measure the performance of the unguided systems, i.e., comparing \mlc, the bare prover and the non-learning \rlc using only UCT with the simple goal-counting proof state evaluation
heuristic. The results of runs with a limit of $200000$ inferences are
shown in Table~\ref{NoLearnM40}. \mlc here performs relatively slightly better than on M2k.
It solves $13450$ problems in total with a higher limit of $2*10^6$
inferences, and  $13952$ problems in total with a limit of $4*10^6$ inferences.

\begin{table}[htbp]
\begin{small}
  \caption{\label{NoLearnM40} Performance on the Miz40 dataset of \mlc, the bare prover and non-learning \rlc with UCT and simple goal-counting proof state evaluation. (200000 inference limit).}
\centering
  \begin{tabular}{llll}
    \toprule
System &    \mlc &   bare prover & \rlc without policy/value (UCT only) \\
Training problems proved & 10438 &  4184 &  7348 \\
Testing problems proved & 1143 &  431 &  804 \\
Total problems proved & 11581 &   4615 &  8152  \\\bottomrule
  \end{tabular}
\end{small}
\end{table}

Finally, we run $10$ iterations of full \rlc with UCT and both policy
and value learning. Only the training set problems are however used
for the policy and value learning.  The inference limit is again
200000, the $0$th run is as above, solving 7348 training and 804
testing problems.  Table~\ref{ResultsTableM40} shows the problems
solved by iterations $1$ to $10$.  \rlc guided by the policy and value
learned on the training data from iterations $0 - 4$ proves (in the
$5$th iteration) $1624$ testing problems, which is $42.1$\% more than
\mlc run with the same inference limit. This is our final result, comparing the baseline prover with the
trained prover on previously unseen data. $42.1$\% is an unusually high improvement which we achieved with practically no domain-specific engineering. Published improvements in the theorem proving field are typically between 3 and 10 \%.
\begin{table}[htbp]
\begin{small}
  \caption{\label{ResultsTableM40}10 iterations of \rlc with policy and value guidance on the Miz40 dataset. Only the training problems are used for the policy and value learning.}
\centering
\setlength\tabcolsep{5pt}
  \begin{tabular}{lllllllllll}
    \toprule
Iteration &     1 &     2 &     3 &     4 &     5 &     6 &     7 &     8 &     9 &    10 \\\midrule
 Training proved &12325&13749&14155&14363&14403&14431&14342&{\bf 14498}&14481&14487\\
Testing proved & 1354 & 1519 & 1566 & 1595 & {\bf 1624} & 1586 & 1582 & 1591 & 1577 & 1621 \\\bottomrule
  \end{tabular}
\end{small}
\end{table}

\subsection{Examples}
\begin{figure}
  \centering
\scalebox{.9}{
  \begin{minipage}{0.6\textwidth}
\begin{tikzpicture}[scale=0.5,sibling distance=2cm,level distance=2cm,
  every node/.style = {shape=rectangle, rounded corners,
    draw, align=center, fill=white}]]
  \tiny
\node [thick] {r=0.3099\\n=1182}
child { node [thick] {p=0.24\\r=0.3501\\n=536 }
  child { node [thin] {p=0.21\\r=0.1859\\n=28\\[-1mm]\large...\vspace{-1.5mm} }}
  child { node [thin] {p=0.10\\r=0.2038\\n=9\\[-1mm]\large...\vspace{-1.5mm} }}
  child { node [thin] {p=0.13\\r=0.2110\\n=14\\[-1mm]\large...\vspace{-1.5mm} }}
  child { node [thin] {p=0.14\\r=0.2384\\n=21\\[-1mm]\large...\vspace{-1.5mm} }}
  child { node [thin] {p=0.14\\r=0.3370\\n=181\\[-1mm]\large...\vspace{-1.5mm} }}
  child { node [thick] {p=0.20\\r=0.3967\\n=279 }
    child { node [thin] {p=0.30\\r=0.1368\\n=14\\[-1mm]\large...\vspace{-1.5mm} }}
    child { node [thin] {p=0.15\\r=0.0288\\n=2\\[-1mm]\large...\vspace{-1.5mm} }}
    child { node [thick] {p=0.56\\r=0.4135\\n=262 }
      child { node [thick] {p=0.66\\r=0.4217\\n=247 }
        child { node [draw=none] {36 more MCTS tree levels until proved}}
      }
      child { node [thin] {p=0.18\\r=0.2633\\n=8\\[-1mm]\large...\vspace{-1.5mm} }}
      child { node [thin] {p=0.17\\r=0.2554\\n=6\\[-1mm]\large...\vspace{-1.5mm} }}
    }
  }
  child { node [thin] {p=0.08\\r=0.1116\\n=3\\[-1mm]\large...\vspace{-1.5mm} }}
}
child { node {p=0.19\\r=0.2289\\n=58\\[-1mm]\large...\vspace{-1.5mm} }}
child { node {p=0.22\\r=0.1783\\n=40\\[-1mm]\large...\vspace{-1.5mm} }}
child { node {p=0.35\\r=0.2889\\n=548\\[-1mm]\large...\vspace{-1.5mm} }};
\end{tikzpicture}
\end{minipage}}
\begin{minipage}{0.39\textwidth}
\begin{scriptsize}
\begin{verbatim}

# (tableau starting atom)


RelStr(c1) 


upper(c1)


Subset(union(c2), carrier(c1)) 


Subset(c2, powerset(carrier(c1))




\end{verbatim}
\end{scriptsize}
\end{minipage}
\caption{\label{mcts}The MCTS tree for the \texttt{WAYBEL\_0:28} problem at the moment when the proof is found.
  For each node we display the predicted probability $p$, the number of visits $n$ and the average reward $r=w/n$. For the
  (thicker) nodes leading to the proof the corresponding local proof goals are presented on the right.}
\end{figure}
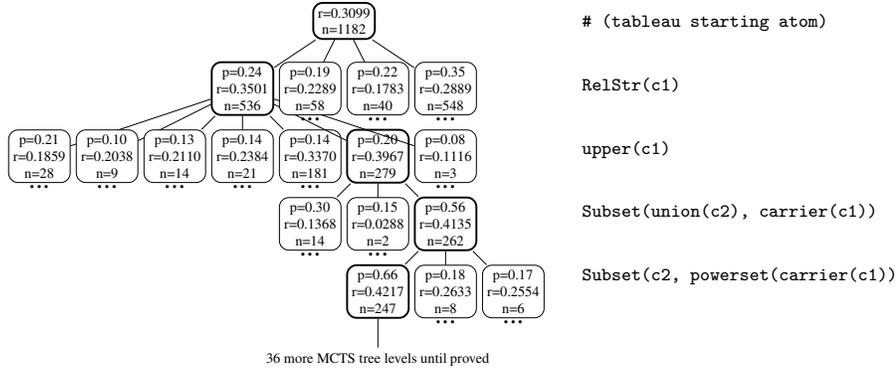

There are 577 test problems that \rlc trained in $10$ iterations on
Miz40 can solve and standard \mlc cannot.\footnote{Since theorem proving is almost never monotonically better, there are
 also 96 problems solved by \mlc and not solved by \rlc in this experiment. The final performance difference between \rlc and \mlc is thus $577 - 96 = 481$ problems.} 
We show three of the problems which cannot be solved by standard \mlc even with a much higher inference limit (4 million). 
Theorem
\texttt{TOPREALC:10}\footnote{\url{http://grid01.ciirc.cvut.cz/~mptp/7.13.01_4.181.1147/html/toprealc\#T10}}
states commutativity of scalar division with squaring for
complex-valued functions.
Theorem \texttt{WAYBEL\_0:28}\footnote{\url{http://grid01.ciirc.cvut.cz/~mptp/7.13.01_4.181.1147/html/waybel_0\#T28}}
states that a union of upper sets in a relation is an upper set. And theorem
\texttt{FUNCOP\_1:34}\footnote{\url{http://grid01.ciirc.cvut.cz/~mptp/7.13.01_4.181.1147/html/funcop_1\#T34}}
states commutativity of two kinds of function composition. All these
theorems have nontrivial human-written formal proof in Mizar, and they
are also relatively hard to prove using state-of-the-art
saturation-style ATPs. Figure~\ref{mcts} partially shows an example of the completed Monte-Carlo tree search for \texttt{WAYBEL\_0:28}. The local goals corresponding to the nodes leading to the proof are printed to the right. 

\begin{lstlisting}[language=Mizar,basicstyle=\ttfamily\scriptsize]
theorem :: TOPREALC:10
for c being complex number for f being complex-valued Function 
 holds (f (/) c) ^2 = (f ^2) (/) (c ^2)

theorem :: WAYBEL_0:28
for L being RelStr for A being Subset-Family of L st 
 ( for X being Subset of L st X in A holds X is upper ) 
 holds union A is upper Subset of L

theorem Th34: :: FUNCOP_1:34
for f, h, F being Function for x being set 
 holds (F [;] (x,f)) * h = F [;] (x,(f * h))
\end{lstlisting}

\section{Related Work}
Several related systems have been mentioned in
Section~\ref{Introduction}.  Many iterations of a feedback loop
between proving and learning have been explored since the
MaLARea~\cite{US+08-long} system, significantly improving over
human-designed heuristics when reasoning in large
theories~\cite{malar14,abs-1802-03375}.  Such systems however only
learn high-level selection of relevant facts from a large knowledge
base, and delegate the internal proof search to standard ATP systems
treated there as black boxes. Related high-level feedback loops have
been designed for invention of targeted strategies of ATP systems~\cite{blistr,JakubuvU17}.

Several systems have been produced recently that use supervised
learning from large proof corpora for guiding the internal proof
search of ATPs. This has been done in the connection tableau
setting~\cite{UrbanVS11,KaliszykU15,FarberKU17}, saturation style
setting~\cite{JakubuvU17a,LoosISK17}, and also as direct automation
inside interactive theorem
provers~\cite{GauthierKU17,GransdenWR15,Whalen16}.
Reinforcement-style feedback loops however have not been explored yet
in this setting.  The closest recent work is~\cite{FarberKU17}, where
Monte-Carlo tree search is added to connection tableau, however without
reinforcement learning iterations, with complete backtracking, and without
learned value. The improvement over the baseline %
measured in that
work is much less significant than here.  An
obvious recent inspiration for this work are the latest reinforcement
learning advances in playing Go and other board games~\cite{SilverHMGSDSAPL16,silver2017mastering,abs-1712-01815,DBLP:conf/nips/AnthonyTB17}.

\section{Conclusion}
In this work we have developed a theorem proving algorithm that uses
practically no domain engineering and instead relies on Monte-Carlo
simulations guided by reinforcement learning from previous proof
searches. We have shown that when trained on a large corpus of general
mathematical problems, the resulting system is more than 40\% stronger
than the baseline system in terms of solving nontrivial new
problems. We believe that this is a landmark in the field of automated
reasoning, demonstrating that building general problem solvers
for mathematics, verification and hard sciences by reinforcement
learning is a very viable approach. 

Obvious future research includes
strong learning algorithms for characterizing mathematical data. We
believe that development of suitable (deep) learning architectures
that capture both syntactic and semantic features of the mathematical
objects will be crucial for training strong assistants for mathematics
and hard science by reinforcement learning.

\bibliographystyle{abbrv}
\bibliography{ate11}

\begin{thebibliography}{10}

\bibitem{abs-1108-3446}
J.~Alama, T.~Heskes, D.~K\"{u}hlwein, E.~Tsivtsivadze, and J.~Urban.
\newblock Premise selection for mathematics by corpus analysis and kernel
  methods.
\newblock {\em J. Autom. Reasoning}, 52(2):191--213, 2014.

\bibitem{IrvingSAECU16}
A.~A. Alemi, F.~Chollet, N.~E{\'{e}}n, G.~Irving, C.~Szegedy, and J.~Urban.
\newblock {DeepMath} - deep sequence models for premise selection.
\newblock In D.~D. Lee, M.~Sugiyama, U.~V. Luxburg, I.~Guyon, and R.~Garnett,
  editors, {\em Advances in Neural Information Processing Systems 29: Annual
  Conference on Neural Information Processing Systems 2016}, pages 2235--2243,
  2016.

\bibitem{DBLP:conf/nips/AnthonyTB17}
T.~Anthony, Z.~Tian, and D.~Barber.
\newblock Thinking fast and slow with deep learning and tree search.
\newblock In I.~Guyon, U.~von Luxburg, S.~Bengio, H.~M. Wallach, R.~Fergus,
  S.~V.~N. Vishwanathan, and R.~Garnett, editors, {\em Advances in Neural
  Information Processing Systems 30: Annual Conference on Neural Information
  Processing Systems 2017}, pages 5366--5376, 2017.

\bibitem{bachmair1994rewrite}
L.~Bachmair and H.~Ganzinger.
\newblock Rewrite-based equational theorem proving with selection and
  simplification.
\newblock {\em Journal of Logic and Computation}, 4(3):217--247, 1994.

\bibitem{barrett2009satisfiability}
C.~W. Barrett, R.~Sebastiani, S.~A. Seshia, C.~Tinelli, et~al.
\newblock Satisfiability modulo theories.
\newblock {\em Handbook of satisfiability}, 185:825--885, 2009.

\bibitem{BlanchetteGKKU16}
J.~C. Blanchette, D.~Greenaway, C.~Kaliszyk, D.~K{\"{u}}hlwein, and J.~Urban.
\newblock A learning-based fact selector for {Isabelle/HOL}.
\newblock {\em J. Autom. Reasoning}, 57(3):219--244, 2016.

\bibitem{hammers4qed}
J.~C. Blanchette, C.~Kaliszyk, L.~C. Paulson, and J.~Urban.
\newblock Hammering towards {QED}.
\newblock {\em J. Formalized Reasoning}, 9(1):101--148, 2016.

\bibitem{xgboost}
T.~Chen and C.~Guestrin.
\newblock {XGBoost}: A scalable tree boosting system.
\newblock In {\em Proceedings of the 22Nd ACM SIGKDD International Conference
  on Knowledge Discovery and Data Mining}, KDD '16, pages 785--794, New York,
  NY, USA, 2016. ACM.

\bibitem{fan2008liblinear}
R.-E. Fan, K.-W. Chang, C.-J. Hsieh, X.-R. Wang, and C.-J. Lin.
\newblock Liblinear: A library for large linear classification.
\newblock {\em Journal of machine learning research}, 9(Aug):1871--1874, 2008.

\bibitem{FarberKU17}
M.~F{\"{a}}rber, C.~Kaliszyk, and J.~Urban.
\newblock {Monte Carlo} tableau proof search.
\newblock In L.~de~Moura, editor, {\em 26th International Conference on
  Automated Deduction (CADE)}, volume 10395 of {\em LNCS}, pages 563--579.
  Springer, 2017.

\bibitem{hh4h4}
T.~Gauthier and C.~Kaliszyk.
\newblock Premise selection and external provers for {HOL4}.
\newblock In X.~Leroy and A.~Tiu, editors, {\em Proc. of the 4th Conference on
  Certified Programs and Proofs (CPP'15)}, pages 49--57. {ACM}, 2015.

\bibitem{GauthierKU17}
T.~Gauthier, C.~Kaliszyk, and J.~Urban.
\newblock {TacticToe}: Learning to reason with {HOL4} tactics.
\newblock In T.~Eiter and D.~Sands, editors, {\em 21st International Conference
  on Logic for Programming, Artificial Intelligence and Reasoning, LPAR-21},
  volume~46 of {\em EPiC Series in Computing}, pages 125--143. EasyChair, 2017.

\bibitem{mizar-in-a-nutshell}
A.~Grabowski, A.~Korni{\l}owicz, and A.~Naumowicz.
\newblock {M}izar in a nutshell.
\newblock {\em J. Formalized Reasoning}, 3(2):153--245, 2010.

\bibitem{GransdenWR15}
T.~Gransden, N.~Walkinshaw, and R.~Raman.
\newblock {SEPIA:} search for proofs using inferred automata.
\newblock In {\em Automated Deduction - {CADE-25} - 25th International
  Conference on Automated Deduction, Berlin, Germany, August 1-7, 2015,
  Proceedings}, pages 246--255, 2015.

\bibitem{DBLP:books/el/RV01/Hahnle01}
R.~H{\"{a}}hnle.
\newblock Tableaux and related methods.
\newblock In Robinson and Voronkov \cite{DBLP:books/el/RobinsonV01}, pages
  100--178.

\bibitem{JakubuvU17}
J.~Jakubuv and J.~Urban.
\newblock {BliStrTune}: hierarchical invention of theorem proving strategies.
\newblock In Y.~Bertot and V.~Vafeiadis, editors, {\em Proceedings of the 6th
  {ACM} {SIGPLAN} Conference on Certified Programs and Proofs, {CPP} 2017},
  pages 43--52. {ACM}, 2017.

\bibitem{JakubuvU17a}
J.~Jakubuv and J.~Urban.
\newblock {ENIGMA:} efficient learning-based inference guiding machine.
\newblock In H.~Geuvers, M.~England, O.~Hasan, F.~Rabe, and O.~Teschke,
  editors, {\em Intelligent Computer Mathematics - 10th International
  Conference, {CICM} 2017}, volume 10383 of {\em Lecture Notes in Computer
  Science}, pages 292--302. Springer, 2017.

\bibitem{holyhammer}
C.~Kaliszyk and J.~Urban.
\newblock Learning-assisted automated reasoning with {F}lyspeck.
\newblock {\em J. Autom. Reasoning}, 53(2):173--213, 2014.

\bibitem{KaliszykU15}
C.~Kaliszyk and J.~Urban.
\newblock {FEMaLeCoP}: Fairly efficient machine learning connection prover.
\newblock In M.~Davis, A.~Fehnker, A.~McIver, and A.~Voronkov, editors, {\em
  Logic for Programming, Artificial Intelligence, and Reasoning - 20th
  International Conference}, volume 9450 of {\em Lecture Notes in Computer
  Science}, pages 88--96. Springer, 2015.

\bibitem{KaliszykU13b}
C.~Kaliszyk and J.~Urban.
\newblock {MizAR 40 for Mizar 40}.
\newblock {\em J. Autom. Reasoning}, 55(3):245--256, 2015.

\bibitem{malar14}
C.~Kaliszyk, J.~Urban, and J.~Vysko\v{c}il.
\newblock Machine learner for automated reasoning 0.4 and 0.5.
\newblock In S.~Schulz, L.~de~Moura, and B.~Konev, editors, {\em 4th Workshop
  on Practical Aspects of Automated Reasoning, PAAR@IJCAR 2014, Vienna,
  Austria, 2014}, volume~31 of {\em EPiC Series in Computing}, pages 60--66.
  EasyChair, 2014.

\bibitem{ckjujv-cpp15}
C.~Kaliszyk, J.~Urban, and J.~Vysko\v{c}il.
\newblock Certified connection tableaux proofs for {HOL L}ight and {TPTP}.
\newblock In X.~Leroy and A.~Tiu, editors, {\em Proc. of the 4th Conference on
  Certified Programs and Proofs (CPP'15)}, pages 59--66. {ACM}, 2015.

\bibitem{DBLP:conf/ecml/KocsisS06}
L.~Kocsis and C.~Szepesv{\'{a}}ri.
\newblock Bandit based monte-carlo planning.
\newblock In J.~F{\"{u}}rnkranz, T.~Scheffer, and M.~Spiliopoulou, editors,
  {\em Machine Learning: {ECML} 2006, 17th European Conference on Machine
  Learning}, volume 4212 of {\em LNCS}, pages 282--293. Springer, 2006.

\bibitem{KuhlweinLTUH12-long}
D.~K{\"u}hlwein, T.~van Laarhoven, E.~Tsivtsivadze, J.~Urban, and T.~Heskes.
\newblock Overview and evaluation of premise selection techniques for large
  theory mathematics.
\newblock In B.~Gramlich, D.~Miller, and U.~Sattler, editors, {\em IJCAR},
  volume 7364 of {\em LNCS}, pages 378--392. Springer, 2012.

\bibitem{Letz1994ControlledIO}
R.~Letz, K.~Mayr, and C.~Goller.
\newblock Controlled integration of the cut rule into connection tableau
  calculi.
\newblock {\em Journal of Automated Reasoning}, 13:297--337, 1994.

\bibitem{LoosISK17}
S.~M. Loos, G.~Irving, C.~Szegedy, and C.~Kaliszyk.
\newblock Deep network guided proof search.
\newblock In T.~Eiter and D.~Sands, editors, {\em 21st International Conference
  on Logic for Programming, Artificial Intelligence and Reasoning, LPAR-21},
  volume~46 of {\em EPiC Series in Computing}, pages 85--105. EasyChair, 2017.

\bibitem{mccune1990otter}
W.~McCune.
\newblock Otter 2.0.
\newblock In {\em International Conference on Automated Deduction}, pages
  663--664. Springer, 1990.

\bibitem{MengP08}
J.~Meng and L.~C. Paulson.
\newblock Translating higher-order clauses to first-order clauses.
\newblock {\em J. Autom. Reasoning}, 40(1):35--60, 2008.

\bibitem{DBLP:journals/aicom/Otten10}
J.~Otten.
\newblock Restricting backtracking in connection calculi.
\newblock {\em {AI} Commun.}, 23(2-3):159--182, 2010.

\bibitem{OB03}
J.~Otten and W.~Bibel.
\newblock {leanCoP:} lean connection-based theorem proving.
\newblock {\em J. Symb. Comput.}, 36(1-2):139--161, 2003.

\bibitem{scikit-learn}
F.~Pedregosa, G.~Varoquaux, A.~Gramfort, V.~Michel, B.~Thirion, O.~Grisel,
  M.~Blondel, P.~Prettenhofer, R.~Weiss, V.~Dubourg, J.~Vanderplas, A.~Passos,
  D.~Cournapeau, M.~Brucher, M.~Perrot, and E.~Duchesnay.
\newblock Scikit-learn: Machine learning in {P}ython.
\newblock {\em Journal of Machine Learning Research}, 12:2825--2830, 2011.

\bibitem{abs-1802-03375}
B.~Piotrowski and J.~Urban.
\newblock {ATPboost}: Learning premise selection in binary setting with {ATP}
  feedback.
\newblock {\em CoRR}, abs/1802.03375, 2018.

\bibitem{robinson1965machine}
J.~A. Robinson.
\newblock A machine-oriented logic based on the resolution principle.
\newblock {\em Journal of the ACM (JACM)}, 12(1):23--41, 1965.

\bibitem{DBLP:books/el/RobinsonV01}
J.~A. Robinson and A.~Voronkov, editors.
\newblock {\em Handbook of Automated Reasoning (in 2 volumes)}.
\newblock Elsevier and {MIT} Press, 2001.

\bibitem{SchaferS15}
S.~Sch{\"{a}}fer and S.~Schulz.
\newblock Breeding theorem proving heuristics with genetic algorithms.
\newblock In G.~Gottlob, G.~Sutcliffe, and A.~Voronkov, editors, {\em Global
  Conference on Artificial Intelligence, {GCAI} 2015}, volume~36 of {\em EPiC
  Series in Computing}, pages 263--274. EasyChair, 2015.

\bibitem{SilverHMGSDSAPL16}
D.~Silver, A.~Huang, C.~J. Maddison, A.~Guez, L.~Sifre, G.~van~den Driessche,
  J.~Schrittwieser, I.~Antonoglou, V.~Panneershelvam, M.~Lanctot, S.~Dieleman,
  D.~Grewe, J.~Nham, N.~Kalchbrenner, I.~Sutskever, T.~P. Lillicrap, M.~Leach,
  K.~Kavukcuoglu, T.~Graepel, and D.~Hassabis.
\newblock Mastering the game of {Go} with deep neural networks and tree search.
\newblock {\em Nature}, 529(7587):484--489, 2016.

\bibitem{abs-1712-01815}
D.~Silver, T.~Hubert, J.~Schrittwieser, I.~Antonoglou, M.~Lai, A.~Guez,
  M.~Lanctot, L.~Sifre, D.~Kumaran, T.~Graepel, T.~P. Lillicrap, K.~Simonyan,
  and D.~Hassabis.
\newblock Mastering chess and shogi by self-play with a general reinforcement
  learning algorithm.
\newblock {\em CoRR}, abs/1712.01815, 2017.

\bibitem{silver2017mastering}
D.~Silver, J.~Schrittwieser, K.~Simonyan, I.~Antonoglou, A.~Huang, A.~Guez,
  T.~Hubert, L.~Baker, M.~Lai, A.~Bolton, et~al.
\newblock Mastering the game of go without human knowledge.
\newblock {\em Nature}, 550(7676):354, 2017.

\bibitem{sutton1998reinforcement}
R.~S. Sutton and A.~G. Barto.
\newblock {\em Reinforcement learning: An introduction}, volume~1.
\newblock Cambridge Univ Press, 1998.

\bibitem{Urban06}
J.~Urban.
\newblock {MPTP} 0.2: Design, implementation, and initial experiments.
\newblock {\em J. Autom. Reasoning}, 37(1-2):21--43, 2006.

\bibitem{blistr}
J.~Urban.
\newblock {BliStr: The Blind Strategymaker}.
\newblock In G.~Gottlob, G.~Sutcliffe, and A.~Voronkov, editors, {\em Global
  Conference on Artificial Intelligence, {GCAI} 2015}, volume~36 of {\em EPiC
  Series in Computing}, pages 312--319. EasyChair, 2015.

\bibitem{US+08-long}
J.~Urban, G.~Sutcliffe, P.~Pudl{\'a}k, and J.~Vysko\v{c}il.
\newblock {MaLARea SG1 - Machine Learner for Automated Reasoning with Semantic
  Guidance}.
\newblock In A.~Armando, P.~Baumgartner, and G.~Dowek, editors, {\em IJCAR},
  volume 5195 of {\em LNCS}, pages 441--456. Springer, 2008.

\bibitem{UrbanVS11}
J.~Urban, J.~Vysko\v{c}il, and P.~\v{S}t\v{e}p{\'a}nek.
\newblock {MaLeCoP}: Machine learning connection prover.
\newblock In K.~Br{\"u}nnler and G.~Metcalfe, editors, {\em TABLEAUX}, volume
  6793 of {\em LNCS}, pages 263--277. Springer, 2011.

\bibitem{Whalen16}
D.~Whalen.
\newblock Holophrasm: a neural automated theorem prover for higher-order logic.
\newblock {\em CoRR}, abs/1608.02644, 2016.

\end{thebibliography}
\end{document}